\pdfoutput=1

\documentclass[11pt]{article}

\usepackage{acl}

\usepackage{times}
\usepackage{latexsym}

\usepackage[T1]{fontenc}

\usepackage[utf8]{inputenc}

\usepackage{microtype}

\usepackage{inconsolata}

\usepackage{graphicx}
\usepackage{amsmath}
\usepackage{amssymb}
\usepackage{booktabs}
\usepackage[normalem]{ulem}

\usepackage{enumitem}
\usepackage{lipsum}
\usepackage{float}

\newcommand{\model}{MAGiC}
\newcommand{\modelfull}{Multi-view Approach to Grounding in Context}

\usepackage{pifont}
\newcommand{\goodcheck}{\color{blue}{\ding{51}}}

\definecolor{mypurple}{RGB}{150, 0, 255}

\newcommand{\xmark}{\color{red}{\ding{55}}}

\newcommand{\minisection}[1]{\noindent{\textbf{#1}.}}
\newcommand\freefootnote[1]{%
  \let\svthefootnote\thefootnote
  \let\thefootnote\relax
  \footnotetext{\textsuperscript{*}#1}
  \let\thefootnote\svthefootnote
}

\definecolor{Gray}{gray}{0.9}

\title{Which One? Leveraging Context Between Objects and Multiple Views for Language Grounding}

\author{Chancharik Mitra\textsuperscript{1*} \quad Abrar Anwar\textsuperscript{2*} \quad Rodolfo Corona\textsuperscript{1} \\
\textbf{Dan Klein\textsuperscript{1}} \quad \textbf{Trevor Darrell\textsuperscript{1}} \quad \textbf{Jesse Thomason\textsuperscript{2}}
\\
\textsuperscript{1}University of California, Berkeley \quad \textsuperscript{2} University of Southern California
}

\begin{document}
\maketitle
\begin{abstract}


When connecting objects and their language referents in an embodied 3D environment, it is important to note that: (1) an object can be better characterized by leveraging comparative information between itself and other objects, and (2) an object's appearance can vary with camera position.
As such, we present the {\modelfull} ({\model}) model, which selects an object referent based on language that distinguishes between two similar objects.
By pragmatically reasoning over both objects and across multiple views of those objects, {\model} improves over the state-of-the-art model on the SNARE object reference task with a relative error reduction of 12.9\% (representing an absolute improvement of 2.7\%).
Ablation studies show that reasoning jointly over object referent candidates and multiple views of each object both contribute to improved accuracy. Code: \url{https://github.com/rcorona/magic_snare/}
\end{abstract}
\freefootnote{Denotes Equal Contribution}
\section{Introduction}\label{sec:introduction}

To distinguish a ``thin handled mug'' between two mugs, we must contextually reason about the object with the \textit{relatively thinner} handle. 
Such \textit{grounded language} can connect to machine representations of the world~\cite{Harnad_1990}.
Considering pragmatic context~\cite{potts_157pragmatics_2022, fried_pragmatics_2022} in grounded natural language can assist applications in vision and robotics~\cite{tellex2020robots, krishna2017visual, lu_vilbert_2019, li_grounded_2022, desai_virtex_2021}.
Additionally, object features like mug handles may be occluded from certain viewpoints, requiring multiple views or 3D information~\cite{huang_multi-view_2022, Wang2021DETR3D3O}.

\begin{figure}[th]
    \centering
    \includegraphics[width=1.0\linewidth]{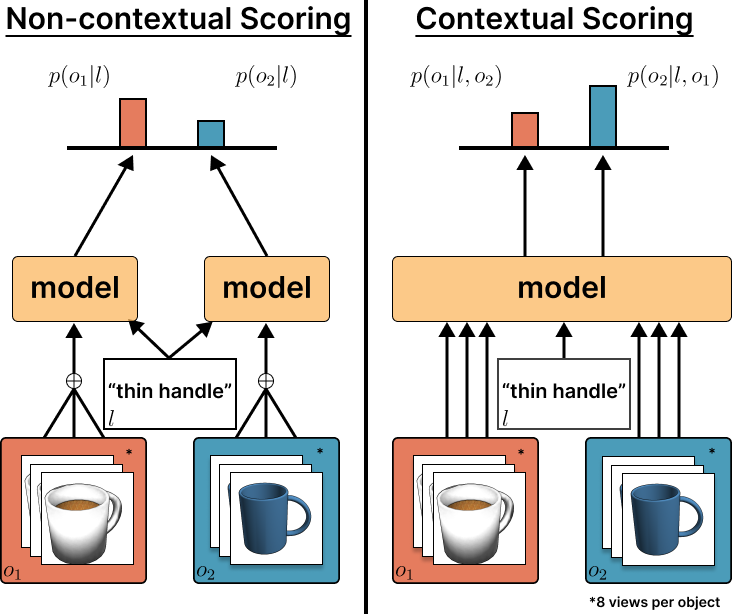}
    \caption{
    \textit{Left:} Previous methods for identifying object referents of language expressions in the SNARE benchmark consider target and distractor objects independently and pool multiple views before grounding.
    \textit{Right:} By contrast, {\model} jointly reasons over target and distractor objects and their views from different angles to identify the correct referent with higher accuracy than the previous state-of-the-art model.
    }
    \label{fig:overview}
\end{figure}

In the real world, language use is situated in a 3D environment and must consider a rich context of alternatives.
However, for tasks like object disambiguation, some models score referring expression compatibility with visual observations of an object in isolation~\cite{thomason_language_2021,corona_voxel-informed_2022}, while broader methods for aligning vision and language representations often consider only static images of objects and scenes~\cite{radford_learning_2021,kim2021vilt}.
Such work can miss language information which can contain comparative information in language and embodied visual information from multiple viewpoints.

We introduce {\modelfull} ({\model}) 
\model\ jointly reasons over candidate referent objects \textit{and} considers each object from multiple possible vantage points (Figure~\ref{fig:overview}). 
We evaluate \model\ via the \textbf{S}hape\textbf{N}et \textbf{A}nnotated with \textbf{R}eferring \textbf{E}xpressions (SNARE) benchmark~\cite{thomason_language_2021}.
In SNARE, candidate objects are always of the same high-level category, such as \textit{chair} or \textit{mug}, and language references uniquely identify one target referent object in contrast to the distractor object of the same category.
Embodied agents operating in real-world environments analogously need to disambiguate between similar objects, such as mugs in a kitchen, parts on a conveyor belt, or rocks on the seafloor.
By reasoning about both objects and their views, \model\ achieves a relative error reduction of 12.9\% (improved accuracy by 2.7\%).
Our contributions include:
\begin{itemize}[noitemsep,nosep]
    \item {\model}, a transformer-based model that reasons over multiple 2D-image views of 3D objects and implicitly considers the relative differences between objects;
    \item state-of-the-art SNARE accuracy;\footnote{\href{https://github.com/snaredataset/snare\#leaderboard}{https://github.com/snaredataset/snare\#leaderboard}}
    \item ablation studies that show both multi-object and multi-view inputs are needed for the {\model} accuracy gains; and
    \item analysis showing {\model} outperforms previous methods even with fewer available object viewpoints.
\end{itemize}


\begin{figure*}[th]
    \centering
    \includegraphics[width=0.9\textwidth]{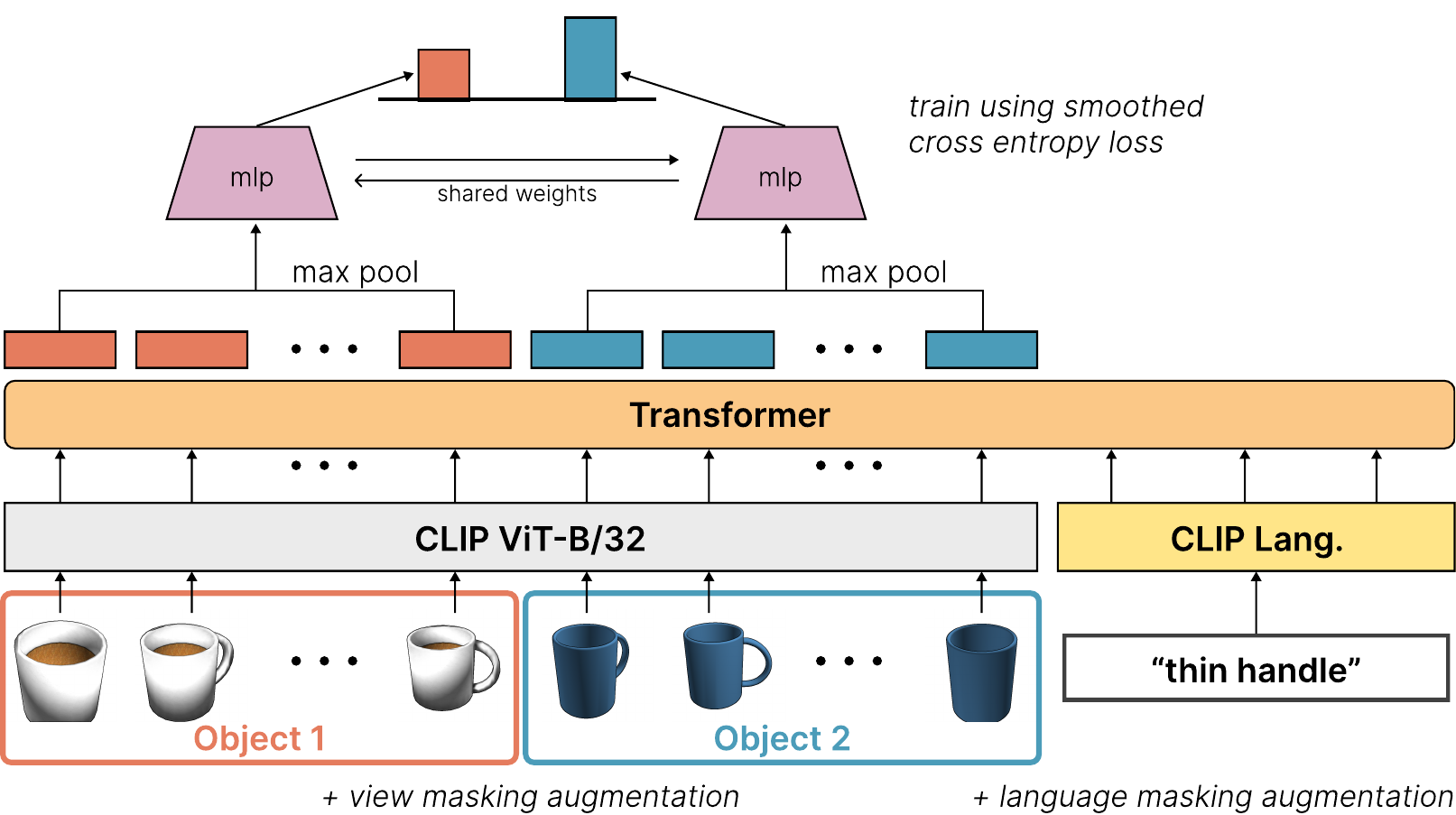}
    \caption{\textbf{Model Architecture.} 
        {\model} consists of a multi-view transformer that attends to CLIP language embeddings for the description and CLIP image embeddings across multiple views for both objects. 
        This transformer allows our model to contextually reason across views about both objects at the same time with respect to a language description. 
        We do not use any positional encodings, and {\model} is invariant to the input order of images and objects.
        Unlike previous methods for SNARE, we pool information from object views only after updating their representations with respect to the language referring expression.
        We apply view masking and language masking augmentations to regularize the model during training.}
    \label{fig:model}
\end{figure*}
\section{Background and Related Work}\label{sec:related_work}


Embodied agents increasingly operating alongside humans must understand the relationships between natural language and the objects they reference. To best capture these relationships, our method synthesizes the comparative context afforded by reasoning over multiple objects and considering each one in multiple views.
\subsection{Object Reference Grounding}

Object referent identification selects specific object referents given natural language descriptors.
Several datasets are prominent in 3D object referent identification.
ShapeGlot \cite{achlioptas_shapeglot_2019} focuses on chairs and lamps, training models to distinguish target objects using shape-based descriptions.
PartGlot \cite{koo2022partglot} employs a reference task for implicit learning of point cloud part-segmentation.
SNARE \cite{thomason_language_2021} uses the ShapeNetSem dataset, featuring 262 object categories, while ShapeTalk \cite{achlioptas2022changeIt3D} introduces 29 object classes for learning grounded point cloud representations.
We utilize the SNARE dataset, leveraging extensive object variety to highlight generalizability.

Previous SNARE task methods scored objects individually \cite{thomason_language_2021, corona_voxel-informed_2022}.
We discuss the limitations of these methods by considering two specific principles in pragmatics \cite{potts_157pragmatics_2022}.
The first is the consideration of contrastive object sets in reference games~\cite{Andreas2016ReasoningAP, bao_learning_2022}. 
Another relevant pragmatics principle relevant to our work is the consideration of alternatives \cite{fried_pragmatics_2022}.
These principles suggest the importance of utilizing comparative information between presented objects when completing SNARE or similar tasks.

\model\ employs language grounding to capture object distinctions in the SNARE task.
Our core insight is joint reasoning over both objects, diverging from methods that independently score reference-referent and reference-distractor pairs~\cite{thomason_language_2021, corona_voxel-informed_2022}, which draws on prior work in pragmatics and reference grounding~\cite{clark1986referring, Vries2016GuessWhatVO, frank_2016, degen2012optimal, franke2014pragmatic, Monroe2017ColorsIC}.

\subsection{3D Language Grounding}

In the domain of grounding language to visual representations, significant progress has been made in 2D \cite{sadhu_zero-shot_2019, yu_modeling_2016, plummer_flickr30k_2016, wang_improving_2021}. 
This research can be extended to work in three dimensions, incorporating more information such as the relative positions and views of multiple objects.
There are many common 3D object representations such as point clouds \cite{qi_pointnet_2017,guo_deep_2020}, meshes \cite{lin2021mesh,bouritsas2019neural}, voxels \cite{yagubbayli2021legoformer}, and neural radiance fields \cite{mildenhall2021nerf,yu_pixelnerf_2021}.

Applications of language and 3D representations include resolving spatial reference for language localizing objects in a 3D scene \cite{zhang2017deepcontext, huang2018cooperative, huang_multi-view_2022}. 
Language guidance can also inform real-world tasks in 3D such as vision-and-language navigation \cite{gu2022vision} or robot instruction following \cite{shridhar_alfred_2020, shridhar2022peract}. 
In all these tasks, grounded language understanding of objects from different viewpoints is necessary. 

The necessity of this 3D, rotational understanding is more prominent in 3D object referent identification tasks such as SNARE \cite{thomason_language_2021} and ShapeGlot \cite{achlioptas_shapeglot_2019}. 
While the model may be presented with explicit 3D object representations to provide rotational information in other identification tasks, SNARE provides multiple 2D views of the referent and distractor objects. 
The previous SoTA methods on SNARE have all aggregated these views before generating a score for each object. 
However, in keeping with Grice's maxim of quantity \cite{grice1975logic}, the \model\ transformer attends over all the views of both objects, in contrast to previous methods attempting SNARE that performed early fusion on view representations.

\definecolor{Gray}{gray}{0.9}

\newcommand*{\belowrulesepcolor}[1]{%
  \noalign{%
    \kern-\belowrulesep 
    \begingroup 
      \color{#1}%
      \hrule height\belowrulesep 
    \endgroup 
  }%
} 
\newcommand*{\aboverulesepcolor}[1]{%
  \noalign{%
    \begingroup 
      \color{#1}%
      \hrule height\aboverulesep 
    \endgroup 
    \kern-\aboverulesep 
  }%
}

\begin{table*}[t]
\centering
\setlength{\aboverulesep}{0pt}
\setlength{\belowrulesep}{0pt}
\resizebox{\textwidth}{!}{
\begin{tabular}{lcc
>{\columncolor{Gray}}r 
>{\columncolor{Gray}}r 
>{\columncolor{Gray}}r
    rrr} 
    & Considers & Lang Attends  & \multicolumn{3}{c}{\textbf{VALIDATION ACC.}} & \multicolumn{3}{c}{\textbf{TEST ACC.}}             \\
Model & Both Objects & to Ind. Views & \multicolumn{1}{l}{\cellcolor{Gray}Visual}      & \multicolumn{1}{l}{\cellcolor{Gray}Blind}       & \multicolumn{1}{l}{\cellcolor{Gray}{All}}         & \multicolumn{1}{l}{Visual}        & \multicolumn{1}{l}{Blind}         & \multicolumn{1}{l}{All} \\
\toprule
Human (U) & \goodcheck & \goodcheck & $94.0\phantom{ (0.0)}$       & $90.6\phantom{ (0.0)}$         & $92.3\phantom{ (0.0)}$         & $93.4\phantom{ (0.0)}$           & $88.9\phantom{ (0.0)}$           & $91.2\phantom{ (0.0)}$           \\ 
\midrule
ViLBERT & \xmark & \goodcheck & $89.5\phantom{ (0.0)}$        & $76.6\phantom{ (0.0)}$         & $83.1\phantom{ (0.0)}$         & $80.2\phantom{ (0.0)}$           & $73.0\phantom{ (0.0)}$           & $76.6\phantom{ (0.0)}$           \\
MATCH   & \xmark & \xmark & $89.2 (0.9)$  & $75.2 (0.7)$  & $82.2 (0.4)$  & $83.9 (0.5)$    & $68.7 (0.9)$    & $76.5 (0.5)$    \\
LAGOR   & \xmark & \xmark & $89.8 (0.4)$  & $75.3 (0.7)$  & $82.6 (0.4)$  & $84.3 (0.4)$    & $69.4 (0.5)$    & $77.0 (0.5)$    \\
VLG     & \xmark & $\sim$ & $91.2 (0.4)$  & $78.4 (0.7)$  & $84.9 (0.4)$  & $86.0\phantom{ (0.0)}$           & $71.7\phantom{ (0.0)}$           & $79.0\phantom{ (0.0)}$           \\ 
{\model}    & \goodcheck & \goodcheck & $\pmb{92.1 (0.4)}$  & $\pmb{81.3 (0.9)}$  & $\pmb{86.8 (0.5)}$  & $\pmb{87.7\phantom{ (0.0)}}$ & $\pmb{75.4\phantom{ (0.0)}}$ & $\pmb{81.7\phantom{ (0.0)}}$ \\ \bottomrule
\end{tabular}}
\caption{\textbf{SNARE Benchmark Performance.} 
    Mean accuracy $\pm$ standard deviation over 10 seeds for existing SNARE approaches, whether those approaches reason over objects jointly, and whether they perform language grounding over individual object views versus pooled representations. 
    Note: $\sim$ indicates that VLG enables language grounding to LegoFormer~\cite{yagubbayli2021legoformer} features of object views, but not RGB views.
    We find that {\model} outperforms all other models on SNARE and is statistically significantly better than VLG, the previous state-of-the-art approach, under a Welch's unpaired two-tailed t-test with a $p<0.001$. 
}
\label{tab:benchmarks}
\end{table*}
\section{Reference Grounding}\label{sec:problem_statement}

We define a reference grounding task where, given one or more visual views of candidate objects and a natural language description, the reference grounding task is to select the object identified by the contrastive referring expression. 
Formally, a model must use a given language description $l$ to predict a target object $o^l$ that is aligned with the language description from among a set of $m$ objects $O = \{o^l, o^{c_1}, o^{c_2}, ..., o^{c_{m-1}}\}$. 
Besides object $o^l$, there are $m-1$ distractor objects $o^{c_i}$ that contribute to the context in which a model needs to reason about.
For each object $o$, the model is able to perceive $n$ views for each object $o_1, ..., o_n$. 
These objects are unordered, and we do not assume access to the relative positions between each view. 
The goal of the task is to learn a classifier function $f(O, l) \rightarrow [0,1]^{m}$ such that a higher probability is assigned to the target object.

Previous approaches \cite{koo2022partglot, achlioptas2022changeIt3D, thomason_language_2021, corona_voxel-informed_2022} learn $f$ for single objects, then each object $o \in O$ is scored separately using a single-object classifier $s(o, l)\rightarrow [0,1]$. 
While classifying only individual objects simplifies the implementation, it limits the model's ability to comparatively reason about objects in context.
Also, previous image-based methods for reference grounding tasks \cite{thomason_language_2021, corona_voxel-informed_2022} aggregate each object's $n$ views without reasoning about each view's relationship to the language description.
To overcome these limitations, a model needs to address two key challenges: 1) reasoning about the contextual relationships between objects, and 2) reasoning about multiple views of each object in relation to the language description.
We propose {\model}, a transformer-based architecture that enables joint reasoning over object-specific and view-specific contextual dependencies for 3D language grounding.

\section{{\model}}\label{sec:model}

We introduce {\modelfull} ({\model}) for language grounding of 3D objects (Figure~\ref{fig:model}). 
In contrast to previous work that individually score each object, {\model} considers both the language and the objects, along with their views, simultaneously.


The design of our model is guided by principles in 3D language grounding and pragmatics.
In SNARE, a model should consider information about comparative differences between two objects to identify the correct referent of the language expression.
To enable a model to more effectively ground language to these visual dissimilarities, we focus on two key elements of the model formulation: (1) \textbf{object context}, which involves jointly reasoning over both objects and the referring expression, and (2) \textbf{multi-view context}, where multiple views of the object representation are explicitly utilized throughout the model without aggregating their representation as a preprocessing step.
We adopt this paradigm in 3D language grounding and design our model to concurrently process features from \textit{both} objects and the referring expression to leverage context-dependent information. 

With the context established by considering both objects and the referring expression, our model can leverage context-dependent information effectively. 
More concretely, consider the scenario of the model being asked to choose between two chairs given the referring expression ``the tall, skinny chair". 
The model can exploit context-dependent information, such as using the descriptor ``tall" to reason over both objects comparatively to ascertain which is taller. 
Additionally, by incorporating features from multiple views of the object, our model benefits from the additional 3D perspective, ensuring that important object information, even if initially rotated out of view, is captured and utilized.


A transformer architecture is well-suited for context-based 3D language grounding due to its wide receptive field and low inductive bias \cite{vaswani2017attention}. 
Unlike CNN-based architectures that have a spatial locality bias, transformers have a wide receptive field that includes \textit{all} input features after just one transformer layer. 
This architecture enables our model to attend to all inputs and effectively leverage both object and multi-view context for 3D language grounding.
Moreover, the low inductive bias of transformers makes the design choice suitable for 3D language grounding, as the transformers are particularly good at handling multiple modalities \cite{xu_multimodal_2023}.

\subsection{Model Architecture}

Given a target object $o^l$, a single distractor object $o^c$, and the language description $l$, {\model} employs a transformer-based architecture to learn a classifier $f([o^l, o^c], l)$. 
We conjecture that our architecture will effectively learn contextual relationships between views and objects.
Our approach focuses on the use of images from each view to represent an object, without relying on additional depth or camera information.
Thus, each object $o$ has $n$ views that represent the object.
Unlike previous work that used additional features, such as voxel-based information~\cite{corona_voxel-informed_2022} or point cloud information \cite{huang_multi-view_2022,achlioptas_shapeglot_2019}, we demonstrate the effectiveness of using image-based views alone for 3D language grounding.
Thus, our model is agnostic to specific orderings of views for an object.

For each view, we utilize a CLIP-ViT~\cite{radford_learning_2021} image encoder $g$ to obtain view-specific visual embeddings $v_i = g(o_i)$. 
Similarly, a CLIP language encoder $h$ is employed to encode the given language description $l$, generating a sequence of token embeddings $[e_d^1, ..., e_d^k] = h(l)$. 
Similar to the previous state-of-the-art model \cite{corona_voxel-informed_2022}, we use the token-level text embeddings from CLIP rather than the CLIP's end-of-token feature that SNARE's baselines use \cite{thomason_language_2021}. 
To distinguish between image-view embeddings and language embeddings, we add a learned token-type embedding to each token to indicate whether it is an image-view embedding or a language embedding \cite{kim2021vilt}.
To ensure permutation invariance between objects, we do not add a token embedding to distinguish whether a view belongs to the first or second object.
To remain agnostic to view orderings, we exclude positional encodings from all views.

Using these representations for the objects and language, we construct a sequence $r=[v_0^l, ..., v_n^l, v_0^c, ... v_n^c, e_l^1, ..., e_l^k]$, which is then passed as input to the transformer encoder $t$:
$$[w_0^l, ..., w_n^l, w_0^c, ... w_n^c, q_l^1, ..., q_l^k] = t(r),$$
where $w$ is a contextualized representation for an object's view, and $q$ are output representations for the language input.
The resulting contextualized representations capture the interplay between views and the language input. 

The object-specific output representations $w_0, ..., w_n$ from the transformer $t$ for an object $o$ are aggregated using max pooling, yielding a single aggregate embedding $u$ representing object $o$.
This aggregate embedding captures the contextual relationships between multiple views of the object in consideration. 
A classifier MLP $s(u)$ takes the contextualized embeddings for an object $o$ as input and generates a score $s$ indicating the likelihood of the object being the target. 

Given a target object $o^l$ that is aligned with a language description $l$ and a single distractor object $o^c$, we apply a sigmoid to the scores for each object to compute the probabilities $p(o^l|l,o^c)$ and $p(o^c|l,o^l)$ of the target and distractor objects, respectively.




\subsection{Attention Masking Augmentation}

Humans often adapt and rely on a subset of views or language cues when faced with challenging circumstances or limited information. 
This observation motivates the exploration of masking techniques in language grounding tasks, aiming to enhance model performance by selectively blocking out certain inputs and encouraging the model to focus on the most relevant information.

We incorporate attention masking augmentations into our model, specifically targeting the transformer's attention weights for both the view and language inputs \cite{girdhar2021anticipative, vaswani2017attention, cho2022cross}. This masking strategy encourages the model to develop a better understanding of multi-view contextual relationships and effectively capture the essential aspects for accurate predictions.

For view masking, we introduce a 10\% probability of masking out each individual view during training.  
This process promotes view invariance as well as the ability to generalize to unseen viewpoints.
Similarly, for language masking, we apply a 20\% probability of masking out each word in the input language description. 
By randomly masking a portion of the word and image embeddings, we encourage the model to learn more robust vision and language representations that are capable of handling missing or incomplete information.


\begin{table}[t]
\setlength{\aboverulesep}{0pt}
\setlength{\belowrulesep}{0pt}
\centering
\resizebox{\columnwidth}{!}{\begin{tabular}{@{}l
>{\columncolor{Gray}}r
>{\columncolor{Gray}}r
>{\columncolor{Gray}}r
}
& \multicolumn{3}{c}{\textbf{VALIDATION ACC.}} \\
Model                     & \multicolumn{1}{l}{\cellcolor{Gray}Visual}              & \multicolumn{1}{l}{\cellcolor{Gray}Blind}               & \multicolumn{1}{l}{\cellcolor{Gray}All}                 \\
\toprule
MATCH                     & $90.6 (0.5)$          & $77.0 (0.7)$          & $83.9 (0.4)$          \\
\quad+ obj. context            & $90.5 (0.5)$          & $76.8 (0.6)$          & $83.7 (0.3)$          \\ \midrule
{\model} & $\pmb{92.1 (0.4)}$ & $\pmb{81.3 (0.9)}$ & $\pmb{86.8 (0.5)}$ \\
\quad- obj. context            & $91.1 (0.5)$          & $79.4 (1.1)$          & $85.3 (0.5)$          \\
\quad- mv. context             & $91.0 (0.6)$          & $79.5 (0.8)$          & $85.3 (0.4)$          \\
\quad- both contexts           & $90.5 (0.6)$          & $78.2 (1.2)$          & $84.4 (0.6)$          \\ \bottomrule
\end{tabular}}
\caption{\textbf{Context Ablations.} 
We investigate the importance of multi-view context (mv. context) and object context (obj. context). 
On the validation set, we report 10 averaged seeds and the standard deviation on ablations of both contexts for MATCH and {\model}. 
We note that the MATCH performance is different from Table~\ref{tab:benchmarks} as these are our replications of MATCH results as opposed to the paper~\cite{thomason_language_2021} report.
We find that if we remove one type of context or both, performance is degraded for {\model}.
}
\label{tab:context_ablation}
\end{table}

\section{Evaluation}\label{sec:evaluation}

We evaluate the effectiveness of our method on the SNARE~\cite{thomason_language_2021} benchmark, a language grounding task that draws from a subset of items in the ShapeNetSem~\cite{shapenet2015,savva2015semantically} dataset, specifically those included in the ACRONYM~\cite{eppner_acronym_2020} robot grasping dataset. 
The SNARE benchmark adversarially selects similar target and distractor objects to challenge 3D language grounding approaches. 
In the object reference task, the model is presented with a natural language description $l$ and must correctly identify the target object $o^l$ from a set of $m=2$ objects $O = \{o^l, o^{c_1}\}$. 
In Section~\ref{sec:multi-distractor}, we investigate varying $m$.
Each object $o$ in the benchmark is accompanied by $n=8$ image views, capturing the object from different perspectives at 45-degree intervals.
As both target and distractor objects are from the same ShapeNetSem category, the SNARE benchmark aims to evaluate a model's contextual reasoning abilities.

The SNARE benchmark encompasses two types of object descriptions: \textbf{visual} and \textbf{blindfolded}. 
Visual descriptions are generated by annotators who are guided to include the object's name, shape, and color. 
These visual descriptions aim to capture a comprehensive understanding of the object, providing relevant visual cues to guide the grounding process (e.g., ``the red mug"). 
On the other hand, blind descriptions predominantly focus on the object's shape and specific distinguishing attributes, intentionally omitting color and other visual characteristics that might aid identification (e.g., ``the one with a tapered lip").

The SNARE benchmark is split into training, validation, and test splits.
The train/validation/test sets are split over (207 / 7 / 48) ShapeNetSem object categories, containing (6,153 / 371 / 1,357) unique object instances and (39,104 / 2,304 / 8,751) object pairings, each accompanied by a referring expression. 
The validation and test sets include unseen object categories that were not encountered during the model training phase, thus evaluating the generalizability and robustness of different methods.

\subsection{Models} 

We compare the performance of {\model} against several baselines, including the previous state-of-the-art (SOTA. 
We describe these baselines below):

\begin{itemize}[label={}]
    \itemsep0pt
    \item \textbf{Human} accuracy serves as an upper bound for performance. These results are provided from SNARE \cite{thomason_language_2021}. Human performance is determined by evaluating whether the annotators can unanimously identify the corresponding object based on the provided natural language description. 
    
    \item \textbf{MATCH}~\cite{thomason_language_2021} uses CLIP-ViT to encode the views of each object. These encoded views are then max-pooled and concatenated to the language description embedding. Then, an MLP is trained to assign scores to each object independently based on the concatenated representation.
    
    \item \textbf{ViLBERT}~\cite{lu_vilbert_2019,thomason_language_2021} uses 14 views as opposed to the standard 8 views in SNARE. These images are tiled into a single image based on the camera view. ViLBERT then attends to the bounding boxes of each view to provide an image representation that is used in a MATCH model instead of the CLIP-ViT encoder.  
    
    \item \textbf{LAGOR}~\cite{thomason_language_2021} (Language Grounding through Object Rotation) builds upon the MATCH model. LAGOR introduces additional regularization through a view prediction loss on each view. The model is presented with only two random views of each object, and it scores each view individually for language grounding in addition to view prediction. 
    
    \item \textbf{VLG}~\cite{corona_voxel-informed_2022} (Voxel-informed Language Grounding) uses a pretrained LegoFormer~\cite{yagubbayli2021legoformer} model for image-to-voxel map prediction. VLG employs a factorized representation of the predicted voxel map, CLIP image embeddings, and CLIP language embeddings to score an object. By incorporating voxel-based information, the VLG baseline serves as a strong comparison against our model, which suggests an alternative pragmatic approach.
    


\end{itemize}

\subsection{Training Details}
We train {\model} on the SNARE dataset using a smoothed binary cross-entropy loss.
We adopt a similar training strategy as VLG. We train our model for 75 epochs using the AdamW optimizer. 
The learning rate is set to \mbox{1e-3}, and we incorporate a linear learning rate warmup for the first 10,000 steps of training. 
Our model uses 3 transformer encoder layers, 8 attention heads, and a hidden size of 256 for a total of 3.6 million trainable parameters. We train our models with a batch size of 64.
\begin{figure}[th]
    \centering
    \includegraphics[width=1.0\linewidth]{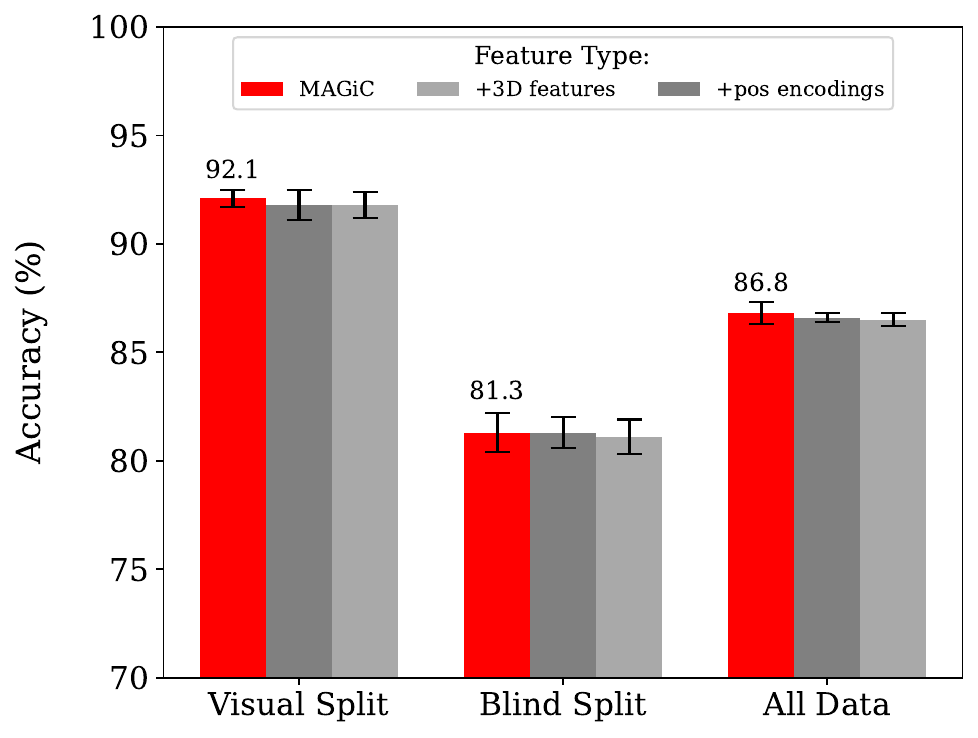}
\caption{\textbf{Explicit 3D Features.} 
    We find that adding 3D structural information to {\model} does not improve accuracy on SNARE.
    }
        \label{fig:feature_additions}
\end{figure}
\section{Results}\label{sec:results}

In this section, we present the test set performance of our model and compare it with the previous state-of-the-art models. 
Additionally, we report the average performance and standard deviation of our model and various ablations on the validation set, calculated over 10 different seeds.

\subsection{{\model} improves over SOTA} 
Table~\ref{tab:benchmarks} presents the performance comparison of models on the SNARE benchmark.
{\model} outperforms all other models with a 2.7\% absolute accuracy improvement on the test set over VLG. 
In the blindfolded split, {\model} has a 3.7\% performance increase over VLG. 
Across the entire validation set, {\model} is statistically significantly better in grounding accuracy than VLG with a $p<0.001$ under a Welch's unpaired two-tailed t-test.


MATCH aggregates the CLIP embeddings using max pooling, removing its ability to effectively reason over the 3D structure of an object.
VLG explicitly uses 3D features and improves 2.5\% on grounding accuracy compared to MATCH.
{\model} however is able to improve performance by 5.2\% over MATCH.
These results suggest that our model does not explicitly require additional 3D structure like VLG.

Though VLG also uses a transformer-based architecture, VLG uses max pooling to aggregate image features before it is input into the transformer model. 
In the blindfolded subset, ViLBERT previously had the top performance of 73.0\%, likely beating VLG since it used 14 views instead of 8 views. 
Although ViLBERT reasons explicitly over multi-view context rather than pooling view information like VLG and MATCH, {\model} improves over ViLBERT by 2.4\% on the blindfolded set using fewer views.
This performance difference implies that by leveraging CLIP image features for each view independently, {\model} demonstrates the ability to capture and reason about multi-view context effectively. 

We believe our performance gain can also be attributed to capturing object and multi-view context. In the next subsection, we present ablations to further demonstrate this result.


\subsection{Ablation Study}

We present several ablations performed on the SNARE validation split. We first investigate the precise contributions of \textbf{object} and \textbf{view} context to our method's improvement on the benchmark as shown in Figure~\ref{tab:context_ablation}. We also examine the effect of additional 3D information and varying the number of views on our method.


\minisection{Context improves validation accuracy} 
In Table~\ref{tab:context_ablation}, we find that using context improves validation accuracy on SNARE, implying that {\model} can capture and utilize contextual dependencies, showcasing its advantage over MLP-based architectures. 
To assess the significance of object context in our model, we added object context to a MATCH model and removed it from {\model}.
We find that adding object context to MATCH does not help improve performance. 
In contrast, removing object context from {\model} decreases grounding accuracy by 1.5\%.
{\model} without object context is similar to the ViLBERT-based MATCH model in Table~\ref{tab:benchmarks}, as both only use multi-view context.
These two models have a noticeable 2.3\% difference in grounding accuracy, though some of this difference could be attributed to ViLBERT's weaker representational capacity for language grounding compared to CLIP.
These results suggest that {\model} is able to effectively leverage object context.

To understand the importance of multi-view context, we remove multi-view context and only reason over object context. 
{\model} without multi-view context is conceptually similar to MATCH with object context, but we find a 1.6\% difference in validation performance between the models.
MATCH's lower performance with multi-view context implies that {\model} can contextually reason between objects better than MLP-based architectures. 

Most notably, we find that {\model} without multi-view and object context has 0.7\% higher overall validation accuracy than MATCH, which is reasonably within error bounds. 
The validation performance of VLG in Table~\ref{tab:benchmarks} also performs similarly to {\model} without both types of context.
The similarity in their performance indicates that the difference between MATCH, VLG, and {\model} comes from {\model}'s ability to reason contextually between both views and objects.


\minisection{{\model} does not require 3D information}
In Figure~\ref{fig:feature_additions},  we investigate whether 3D information is necessary to comparatively ground two objects by conducting two experiments that introduce 3D structure explicitly: via positional encodings and explicitly adding 3D features.


\begin{figure}[t]
    \centering
    \includegraphics[width=0.9\linewidth]{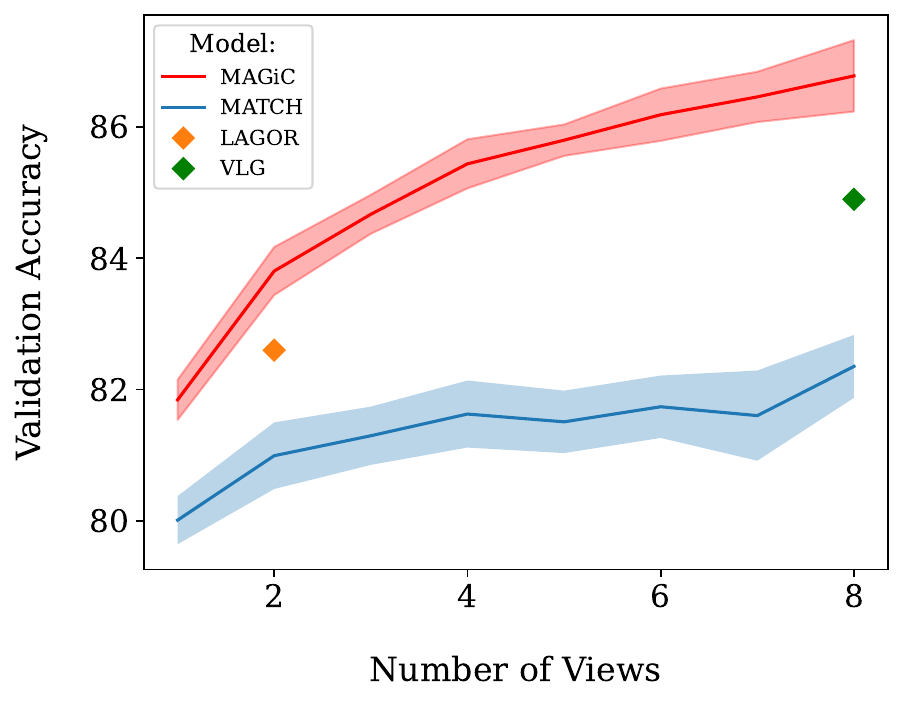}
    \caption{\textbf{Fewer Views Impact on Performance.} We report results on the validation set on the impact of fewer views on performance. We find that {\model} outperforms MATCH, LAGOR, and VLG, achieving greater accuracy with fewer views.}
    \label{fig:few_view}
\end{figure}

\textit{Positional Encodings.} 
Given that our model does not impose any specific ordering for the views, we rely on our model to learn the 3D structure of objects implicitly from unordered 2D-image views.
To investigate the maximum potential of an image-based 3D object grounding model, we experiment with enforcing canonical image view orderings and incorporating learnable positional encodings for the views. 
While {\model} handles unordered input views without relying on knowledge about camera rotations, we specifically enforce a canonical ordering scheme based on 45-degree rotations for the inputs and add learnable positional encodings for each view.
If consistent view orderings and positional encodings help in learning 3D structure, we would expect improved performance. 
However, our findings in Figure~\ref{fig:feature_additions} indicate that enforcing order and using positional encodings do not result in performance changes, implying that {\model} can capture view-specific contextual relationships without explicit positional information.

\textit{3D Features.} The performance gains of VLG over MATCH in Table~\ref{tab:benchmarks} can be attributed to the addition of explicit 3D information.
To assess whether our model can benefit from explicit 3D information, we investigate the impact of incorporating supplementary, view-specific 3D features into the transformer input. 
We use features pre-computed using the Point-E~\cite{nichol_point-e_2022} transformer for each object view and language description. 
Point-E is a language-conditioned point cloud diffusion transformer that captures both 3D and language information through a reconstruction task, so we believe it will effectively capture relevant 3D information.
View masking augmentations are applied as necessary. 
Also, we add token-type embeddings so the model can distinguish between the 2D image features and the 3D features. 
We find that the explicit inclusion of 3D features does not improve accuracy.

These results further reinforce the importance of grounding fine-grained object differences over the use of 3D information in improving comparative language grounding (as posited in prior works).

\minisection{{\model} is more robust to fewer views}
Stronger performance by {\model} on view-limited experiments compared to the previous SOTA demonstrates {\model}'s ability to handle limited visual information in language grounding tasks. 
By retraining {\model} and MATCH on a reduced number of views as shown in Figure~\ref{fig:few_view}, we can assess a model's ability to effectively leverage limited visual information and still accurately understand and interpret natural language descriptions.
We find that on the validation set, {\model} achieves higher accuracy with fewer views compared to other models.
For instance, with only 4 views, {\model} achieves an accuracy of 85.4\%, surpassing VLG, which attains 84.9\% accuracy with 8 views.
This suggests that {\model} can more efficiently leverage available information from fewer views.
Our findings contribute to a deeper understanding of the significance of exploiting multiple views in language-grounding tasks.



\section{Discussion}
In this work, we present {\model}, which demonstrates significant improvements in language grounding accuracy on an object reference task by reasoning jointly over objects and their multi-view contexts when scoring their compatibility with referring expressions. 
We find that comparatively reasoning over multiple objects is central to capturing contextual relationships that enhance the model's ability to ground object descriptions, with added multi-view context also contributing to better language-to-object grounding.
The experimental results from the SNARE object identification benchmark highlight the effectiveness of {\model}, which outperforms all methods on both the validation and test sets. 
\section{Limitations}\label{sec:future}

{\model} heavily relies on having access to multiple views of objects. While using multiple views allows for capturing richer context and improving performance, it also requires obtaining and processing multiple images for each object, which may not always be feasible or practical in certain scenarios. Future work could consider actively selecting views that promote the most information gain. 
Additionally, our experiments focus on a single distractor object. 
We provide preliminary multiple distractor experiments in the appendix to showcase the practicality of \model\ in the real world, which is provides a foundation for future work on comparatively reasoning over multiple objects.

Additionally, {\model} uses CLIP embeddings for encoding visual information. While CLIP provides powerful pre-trained image and text encoders, its representations may not fully capture the intricacies and characteristics of 3D objects. This limitation could potentially impact the model's ability to discriminate between visually similar objects or capture fine-grained details crucial for accurate language grounding.

\section{Potential Negative Societal Impact}

{\model} was designed to ground language to 3D household objects. 
However, {\model} has direct potential uses for sensitive applications such as face identification and surveillance.
For instance, law enforcement agencies may use {\model} with vague witness testimony to discern a suspect given two sets of mugshots with multiple views.
In these high-stakes applications, our model could generate harmful and discriminatory identifications that would further negatively impact historically minoritized peoples.
Furthermore, our model uses a CLIP-backbone, and previous literature has shown that CLIP reinforces malignant sexist and racist stereotypes \cite{hundt2022robots} and exhibits gender bias \cite{wang2022fairclip,agarwal2021evaluating} which are part of broader patterns of marginalization in society. 
Vision-and-language models have also been shown to compound gender biases that exist separately in language and vision \cite{srinivasan2021worst}. 
Therefore, these models must account for the ways in which language and perception reflect social norms.
%

%



\section{Acknowledgements}
The authors would like to thank Jerry A. Yang for helpful discussions on the potential negative societal impact of models like MAGiC. 
RC was funded by the DARPA SemaFor program, BAIR Commons, and an NSF Graduate Research Fellowship.

\bibliography{acl_latex}
\newpage
\appendix

\section{Appendix}
\label{sec:appendix}
In this supplementary section, we describe additional experiments, ablations, and results related to our work.

\subsection{Masking Ablations}

\begin{figure}[th]
    \centering
    \includegraphics[width=1.0\linewidth]{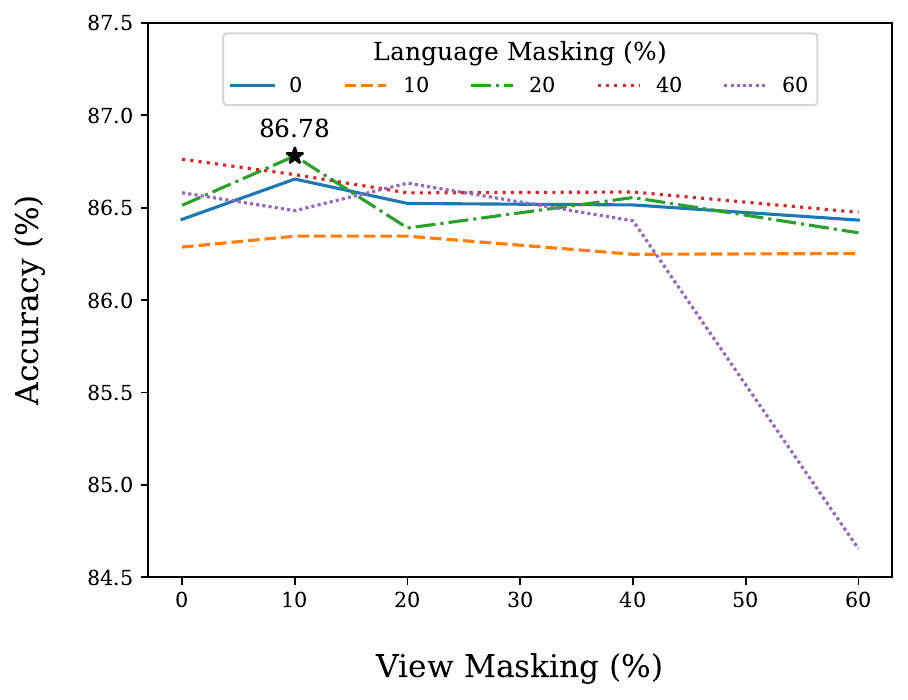}
    \caption{\textbf{View and language masking.} We show the impact of different attention masking percentages for the view and language tokens that are input into \model. Each variant is trained for 10 seeds. We find that 10\% view masking and 20\% language masking achieved the highest validation set accuracy.}
    \label{fig:masking}
\end{figure}

As discussed in Section \ref{sec:model}, in order to improve the robustness and generalization capabilities of \model, we employ masking augmentations on both the language embeddings and the view embeddings as regularization for our model.
Specifically, we applied random masking to a certain percentage of the language and view embeddings during training, analyzing the impact of different masking percentages as depicted in Figure~\ref{fig:masking}. Through hyperparameter tuning on the validation set, we determined that a 20\% language masking and a 10\% view masking yield language grounding accuracy improvements.
We ran each model for 10 seeds.
However, we also noticed that excessive regularization can have a detrimental effect on accuracy, highlighting the need for a balanced application of masking augmentations.
\subsection{Contrastive Loss}

\begin{figure}[th]
    \centering
    \includegraphics[width=1.0\linewidth]{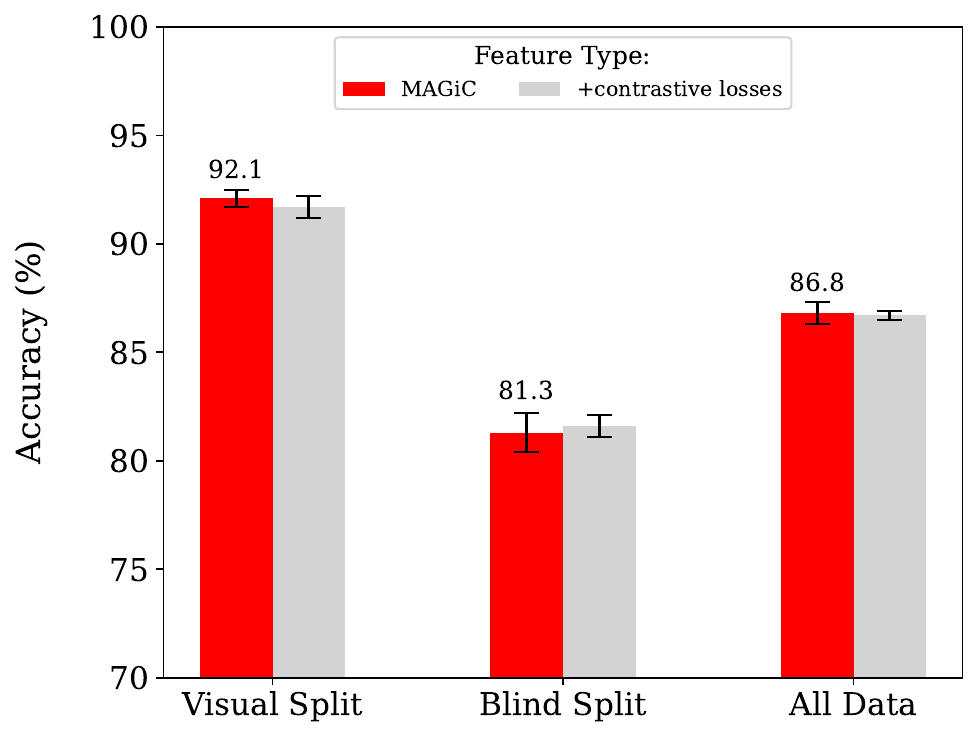}
    \caption{\textbf{Contrastive Loss.} We train {\model} on 10 seeds on the validation set with and without contrastive losses. We show that there is no noticeable impact on \model's validation accuracy when a contrastive loss is added during training.}
    \label{fig:contrastive}
\end{figure}

We investigated additional regularization by using CLIP-like contrastive losses on the output representations. Losses that are similar in spirit have been used in face recognition and clustering research \cite{Schroff_2015} as well as multi-modal sentiment analysis research \cite{hazarika2020misa}. At a high-level, we implement a contrastive loss that motivates the embedded target-object image features to be similar to the embedded object description language features. 
Our model does not have any supervision on the output language representations, and thus, we hypothesized that a contrastive loss would have led to a more structured embedding space. 
Additionally, we expected that the additional supervision from the contrastive loss on the output embeddings from the language inputs would help improve grounding accuracy. 
However, we did not find any improvements on \model's accuracy on the validation set as shown in Figure \ref{fig:contrastive}. 
These findings indicate that the transformer model was already able to contrastively structure the embedding space given access to both objects and the language description such that the added contrastive loss was not further advantageous towards that goal.

\subsection{Additional Discussion}

We would also like to note that our model outperformed another model on the SNARE leaderboard called LOCKET. 
However, we were unable to find any code or paper publicly associated with LOCKET at the time of submission, and omitted it from Table~\ref{tab:benchmarks}.

The code for {\model} will be made public after anonymity restrictions are lifted.

\subsection{Multiple Distractor Experiment}

\label{sec:multi-distractor}
While the SNARE benchmark presents the model with one target and one distractor object, we demonstrate \model's ability to generalize to multiple distractors, as may be the case in a more realistic use case.
SNARE provides adversarially-selected pairs with language annotations.
To have multiple distractor objects, we randomly select an object in the same train/val set.
There is no guarantee for new distractor objects will be in the same category of as the initial two objects since the language might not differentiate additional objects of the same category.

\begin{figure}[th]
    \centering
    \includegraphics[width=1.0\linewidth]{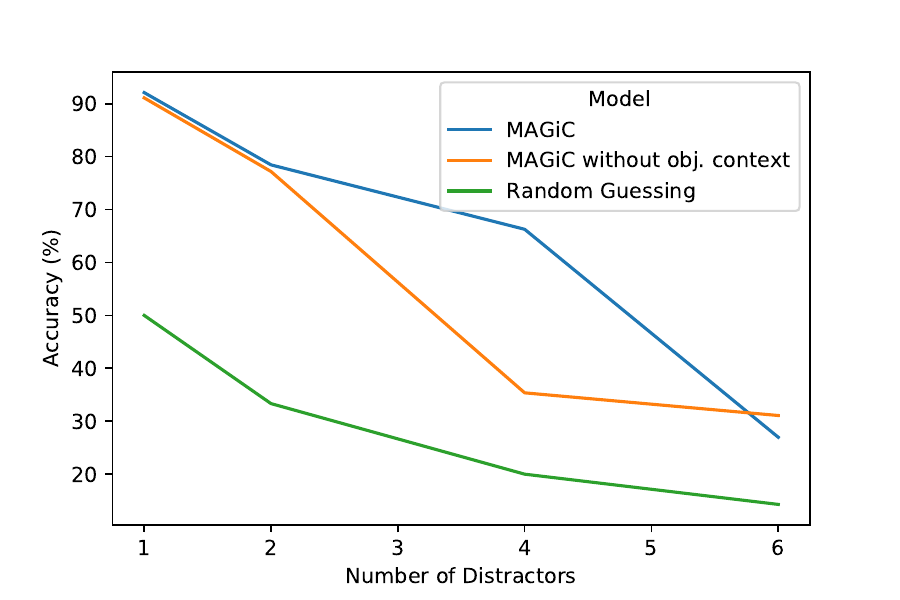}
    \caption{\textbf{Number of distractors.} We show the impact of different numbers of distractors on the performance of {\model} and {\model} without context. Each variant is trained for 1 seed. We find that MAGiC}
    \label{fig:djistractors}
\end{figure}

Our results in Figure~\ref{fig:djistractors} show that while overall performance decreases, \model\ generally retains its strong performance over an architecture without object context.
\model\ without object context is similar to the MLP-based MATCH model, as they score each object individually.
We find that reasoning over all objects generally outperforms scoring the objects individually.
We note that performance clearly degrades as more objects are added, and we show a line depicting random chance to show that our model has generally high performance. 
Due to SNARE being a dataset for loading 2 objects at a time, implementation constraints limited us from scaling up these experiments efficiently. Thus each variant is trained only on 1 seed, which makes it clear that this result becomes noisy as more distractor objects are added. 
We also note that for MAGiC, for additional distractor objects, MAGiC is trained and evaluated on the same number of distractor objects.

\end{document}